\def\year{2022}\relax
\documentclass{article} % DO NOT CHANGE THIS
\usepackage{aaai22}  % DO NOT CHANGE THIS
\usepackage{times}  % DO NOT CHANGE THIS
\usepackage{helvet}  % DO NOT CHANGE THIS
\usepackage{courier}  % DO NOT CHANGE THIS
\usepackage[hyphens]{url}  % DO NOT CHANGE THIS
\usepackage{graphicx} % DO NOT CHANGE THIS
\usepackage{natbib}  % DO NOT CHANGE THIS AND DO NOT ADD ANY OPTIONS TO IT
\usepackage{caption} % DO NOT CHANGE THIS AND DO NOT ADD ANY OPTIONS TO IT
\usepackage{dsfont}
\DeclareCaptionStyle{ruled}{labelfont=normalfont,labelsep=colon,strut=off} % DO NOT CHANGE THIS
\usepackage{algorithm}

\usepackage{newfloat}
\usepackage{listings}
\floatstyle{ruled}
\newfloat{listing}{tb}{lst}{}
\floatname{listing}{Listing}
\usepackage{algorithm}
\usepackage{algpseudocode}
\usepackage{amsmath}
\usepackage{caption}
\usepackage{subcaption}
\usepackage{algpseudocode}

\title{Quantifying Human Bias and Knowledge to guide ML models during Training}
\author {
    Hrishikesh Viswanath,\textsuperscript{\rm 1}
    Andrey Shor, \textsuperscript{\rm 1}
    Yoshimasa Kitaguchi \textsuperscript{\rm 1}
}
\affiliations {
    \textsuperscript{\rm 1} Department of CS, Purdue University\\
    
    hviswan@purdue.edu, ashor@purdue.edu,ykita@purdue.edu
}

\usepackage{bibentry}
\begin{document}

\maketitle

\begin{abstract}
This paper discusses a crowdsourcing based method that we designed to quantify the importance of different attributes of a dataset in determining the outcome of a classification problem. This heuristic, provided by humans acts as the initial weight seed for machine learning models and guides the model towards a better optimal during the gradient descent process. Often times when dealing with data, it is not uncommon to deal with skewed datasets, that over represent items of certain classes, while underrepresenting the rest. Skewed datasets may lead to unforeseen issues with models such as learning a biased function or overfitting. Traditional data augmentation techniques in supervised learning include oversampling and training with synthetic data. We introduce an experimental approach to dealing with such unbalanced datasets by including humans in the training process. We ask humans to rank the importance of features of the dataset, and through rank aggregation, determine the initial weight bias for the model. We show that collective human bias can allow ML models to learn insights about the true population instead of the biased sample. In this paper, we use two rank aggregator methods Kemeny Young and the Markov Chain aggregator to quantify human opinion on importance of features. This work mainly tests the effectiveness of human knowledge on binary classification (Popular vs Not-popular) problems on two ML models: Deep Neural Networks and Support Vector Machines. This approach considers humans as weak learners and relies on aggregation to offset individual biases and domain unfamiliarity.  
\end{abstract}

\section{Introduction}

When dealing with data, a problem that persists is the lack of comprehensive training datasets that are highly representative of the domain. Tablular datasets that represent data points belonging to many classes may not have balanced representation of points of each class. Some target classes may be significantly over represented. When machine learning models are trained on these unbalanced dataset, they may overfit to the majority class, thereby inheriting the bias of the dataset. In order to deal with such issues, data augmentation techniques for such as randomized oversampling and synthetic data generation have been introduced \cite{shorten2019survey}. 
Random oversampling \cite{moreo2016distributional} is a technique in which one randomly selects examples from the minority class, with replacement, and adds them to the training dataset. This approach can be repeated in order to achieve the desired class distribution in the training dataset. The advantages of this approach is its simplicity as well as lack of bias. However, the main disadvantage with oversampling tends to be the fact that since one is reusing existing examples, overfitting is more likely. 
Synthetic data generation is an increasingly popular tool for adding data to datasets with low number of examples to learn from. Synthetic data works by using various algorithms that mirror the statistical properties of the original data \cite{soltana2017synthetic}. One key benefit of synthetic data is that it allows data scientists and engineers to create more examples without violating the confidentiality of the original dataset. This is useful for domains such as healthcare, banking, and even voter turnout. 
The key issue with the second approach is that synthetic data requires the original dataset to be unbiased, as otherwise, it will inherit the bias of the dataset you used, and hence will not help learn a model generalizing to the distribution of which the dataset comes from.

This work proposes an approach to deal with situations when there simply isn't enough data available or is severely skewed and noisy. To mitigate the issue of insufficient training data, we explore a way to model human knowledge as a way to optimize the training process to train the machine learning models to learn a much better function in the hypothesis space. We use human knowledge as initial bias of the weights in an attempt to lead the model towards a better minima during the training process. 
However, the question that arises is the trustworthy nature of human opinion \cite{hilbert2012toward}. A major issue with models that use domain-expert knowledge is that the definition of domain-expert can greatly vary the training process. We need to account for personal biases of humans to ensure that the models themselves don't become biased. 
Our approach doesn't consider human beings as domain-experts but rather as weak, biased learners. This allows us to treat humans as ensemble learners and by collectively aggregating the opinions of multiple humans, we can mitigate personal biases but capture a collective bias held by the society, which, we argue, helps the models learn better. 
For our experiments, the information collected by humans are the factors that are important in classification problems. We present a small subset of training data, in tabular form, to humans. We then provide definitions of the attributes. Since we consider humans as weak learners, we do not present complex datasets covering esoteric topics, but rather a dataset that can be comprehended by general population. Some datasets considered were the Spotify song dataset, the German credit dataset and the superstore sales dataset. 

After collecting human opinions, which in this case, would be a ranking of the features that are highly influential in the final outcome, we aggregate and use this data as our initial weights for the models and in the following sections, we present our findings through metrics such as accuracy, F1 Score and interpretability. 

\section{Literature Review}

\cite{koh2020concept} explore the idea of concept bottleneck models as a way to impart high level domain knowledge into machine learning models through the use of domain experts during training. Their approach was to train a model to extract the most important features. These features were then tweaked by a domain expert to better represent the element. These features or concepts, were then used to train the model for both classification and regression. Our approach is quite similar, but rather than using a domain expert, we propose to use a large number of humans with a surface level understanding of the subject matter. \cite{alexander1994template} define ways to extract rules from weight templates in an effort to increase interpretability of neural networks and determine why the take the decisions that they do. They define a weight template as a parameterized space of weight features. We use a similar template, but present it to humans as opposed to using them for interpreting the neural network.

\cite{franceschet2011pagerank}, in their work talk about the page rank algorithm, used by Google search engine, to collectively aggregate overall ranking of web pages that appear in the search engine feed. They consider the number of incoming edges and outgoing edges to evaluate the 'importance' of a web page. A page that recieves a hyperlink from a site that has high importance, will become more important. We liked this idea of coagulating importance of features through distribution of weights between features and used this as a baseline approach to build a collective ranking of weights from individual rankings provided by the users.\cite{topchy2003combining} discuss ways to combine multiple weak clusterings such that the final clustering is a better partition than the individual ones. They define consensus functions, which is a measure of the number of times two datapoints appeared in the same cluster, as a way to determine whether two points need to be in the same cluster. 

\cite{lin2004merging} explore approaches to collect video data from multiple sources and ways to merge these rank lists efficiently. They argue that the scores from multiple lists may not be comparable to each other and hence propose functions such as Linear Regression as a way to map these local rank lists to a global rank list. \cite{li2019comparative} address the issue of dealing with combining partial rank lists, in the field of genomic studies. They argue that a lot of aggregation techniques are domain specific and methods that perform well in a particular domain may need perform just as well in other domains. We however, deal with this issue as a training optimization problem, where the model only sees an initial set of training weights. The models will however use limited training data to converge to a minima, so rank aggregation in this case will not change much with the domain. 

\cite{parisi2014ranking} discuss methods to combine unlabelled data from multiple classifiers using the method of covariance matrix. They state that the off-diagonal entries form the rank one matrix with unit norm eigenvector and eigenvalue. The labelled instances are then ranked by sorting the entries of the eigenvector v. \cite{quost2009learning} investigate ways to perform supervised training when the data has uncertain labels. They use a variant of the Adaboost algorithm to weigh the importance of the classifiers to build an ensemble classifier. While this approach looked exciting at the beginning and it gave us a way to rank the weights based on how important a human was, we found no ways to quantitatively evaluate relative importances of multiple human users. However, through the use of verfied badges, we may be able to assign higher weights to the opinions of trusted users. 

\cite{diligenti2017integrating} discuss ways to deal with lack of training data by injecting prior knowledge into the neural network. They have used semantic based regularization to represent prior knowledge, as a set of first-order-logic clauses. We argue that constructing first order logic clauses is not a trivial task and would require extensive domain knowledge and hence, the method discussed in their work was not useful in building fast plug-and-play models.

\cite{jacobs2021measurement} define methods to measure fariness from a social perspective. They argue that often times, there are unobservable constructs that are hard to account for and need to be measured through inference from observable quantities. We will attempt to verify whether using human bias will account for unobservable constructs, in our future work. 

\cite{biswas2020machine} performed a croud sourced study to determine areas where biases were prevalent. They concluded that most people focused on training data rather than the models when determining the cause of biases. They also concluded that these biases are not general, like race and gender, but are more specific to the domain under consideration. 

\cite{hu2021architecture} discuss ways to define entanglement and see if there was a possibility to achieve disentanglement in neural networks. We used methods such as integrated gradients and neuron conductance to test for entanglement of features in the neural network.

\section{Method}
\subsection{Dataset}
For our experiments, we attempted to perform binary classification through the initial weight distribution heuristic obtained from humans.The Spotify song dataset by  \cite{spotifysongs} which is a tabular dataset, which classifies a song as being popular or not on Spotify was used because it is balanced, comprehensive and is easy to understand by most people. The dataset is parititoned into decades covering a large number of songs from 1960s to 2010s. Each song is represented by a set of features such as Danceability, Key, tempo, Time Signature etc. We used the 2010s dataset, which is a balanced dataset of 6400 points, of which exactly 50\% are classified as positive, with the rest being negative. Positive denotes that a song was popular and negative indicating that a song was not popular. 

\subsection{Extracting Human Opinions}
Methods like Concept Bottleneck Models \cite{koh2020concept} use intermediate representations that are tuned by domain experts. These intermediate representations do not represent the weights but rather abstract classes. The key issue with this method is practicality. Hiring domain experts to tune the values may prove to be expensive and time consuming. In some of these cases, it is hard for a single individual to determine the absolute importance of certain features, but it is easier to determine the relative ordering of the features. In case of problems that do not require an extensive domain knowledge, and can be understood by the general public, we show that collective opinions collected in a crowd-sourced manner can be leveraged to construct an ordering of the weights. Generally, ML models are presented with weights that are randomly initialized. However, in this case, the model will know the relative ordering of the features, but it will need to learn the absolute importance of individual features. This enables the model to achieve a better optimal due to the prior knowledge it posses about the ordering of features. 

To collect data from humans, We first presented a small subset of the data to a group of volunteers. After showing them the ground truth, we presented the features and asked their opinion on which features they thought were relevant in making the final classification. We also ask them to rank the weights in the decreasing order of importance. The volunteers were presented with 20 rows of the dataset that were randomly sampled. Since humans, unlike ML models, have prior general knowledge, 20 rows were sufficient for them to make the decision. Each volunteer then returned an ordering of the features, which were aggregated using the methods mentioned in the following sections. Aggregation of these rankings in theory works like an ensemble of weak learners, because each human has a personal bias, which they take into account while assigning the weights and providing their interpretations of the classification. However, similar to the Ensemble learning techniques, we can make the assumption that a collection of weak (In our case, biased) learners will in expectation provide a distribution of the weights that doesn't deviate from the actual distribution of weights. We emperically showed that the relative ordering in either case were similar.

Algorithm 1 shows the flow of the process. Some things to be noted here is that some of the features may have negative importance. This implies that datapoints with where this feature takes a large value are more likely to be classified as unpopular. This metric can only be applied to binary classification since in a multi class classification, it is harder to capture these kinds of inverse relationships. Another important step in this algorithm is to scale the weights. Often times, initial weights are small values that follow the Normal distribution. To ensure that the initial weights are small enough for the models to learn correctly, we reduced them into the range -1, 1, while preserving the relative ordering because the absolute values of the weights are not important at this stage. 

\begin{algorithm*}[h!]
\caption{Training with Human Bias}\label{alg:cap}
\begin{algorithmic}
\State $H \gets num\_human\_users$
\State $F \gets num\_features$
\State $w_h  \in \Re^{HxF}$
\State $w_{agg} \gets rankAggregator(w_h)$, $w_{agg}  \in \Re^{1xF}$
\For{\texttt{feature in featureSet}}
    \If{feature has negative importance}
    \State $w_{agg}[i] \gets -w_{agg}[i]$
    \EndIf
\EndFor
\State $newMin \gets -1$
\State $newMax \gets 1$
\State $curMin \gets min(w_{agg})$
\State $curMax \gets max(w_{agg})$
\For{\texttt{w in $w_{agg}$}}
    \State $w_{agg}[i] \gets newMin + (newMax - newMin) * (w_{agg}[i] - curMin)/(curMax-curMin)$
\EndFor
\State $weightsInitial \gets w_{agg}$
\State Train(Model, weightsInitial, skewedDataset)
\end{algorithmic}
\end{algorithm*}

\subsection{Aggregating multiple weight rankings}
\subsubsection{Kemeny Young Aggregator}
The Kemeny-Young is a rank aggregator method that is used to merge multiple rankings. The Kemeny rule states that if two people agree with each other, then they both assign a higher ranking to the same candidate. This is called agreement. The Kemeny Algorithm then maximizes this agreement \cite{dwork2001rank}. This is a condorcet method of aggregating votes, by minimizing the Kendall-Tau distance \cite{conitzer2006improved}. 
Kemeny Young Aggregator has been used in the president election/voting system to combine the votes from multiple voters (V) to select a candidate (C).

We modify the algorithm to apply the same for weight aggregation. The set users are denoted by (U) while individual users are denoted by (u). The individual features denoted by (f). Let w denote the weight vector ranking and (W) be the set of rankings or the ranking profile. We aggregate the ranking by minimizing the Kendall-Tau Distance denoted by  $\delta_{KT}^k (w, W)$ \cite{gilbert2020beyond}

\begin{equation}
\delta_{KT}^k (w, W) = \sum_{w' \in W} \delta_{KT}^k (w, w')\\
\end{equation}
\begin{equation}
\begin{split}
\delta_{KT}^k (w, w') = \sum_{(f,f') \in F^2} disagree_{f,f'} (w,w') \\
*\sum_{i=1}^{k-2} (|below_{f}(w) \cup below_{f'}(w')|)
\end{split}
\end{equation}

In equation (2), $disagree_{f,f'}(w,w') = 1$ if $f \succ_w f'$ and $f' \succ_{w'} f$ and 0 otherwise. $below_c(r) = \{x \in C $ s.t. c $\succ_r x\}$. The runtime of the algorithm is $O(n2^n)$ by using Held-Karp algorithm.

\subsubsection{Markov Chain Aggregator}
Markov Chain aggregator uses the concept of transition matrix to produce an combined ranking \cite{dwork2001rank}. We use the same notation as the previous section, where W represents the weight matrix or the ranking profile. 

\begin{equation}
  T_{i,j} =
    \begin{cases}
      -1 & \text{if $(\sum_{k=0}^{|F|} \mathds{1}(W_{i,k} < W_{j,k})) = 0$}\\
      0 & \text{if $(\sum_{k=0}^{|F|} \mathds{1}(W_{i,k} < W_{j,k})) > 0$}\\
      1 & \text{if $(\sum_{k=0}^{|F|} \mathds{1}(W_{i,k} < W_{j,k})) < 0$}
    \end{cases}       
\end{equation}
Normalization:\\
\begin{equation}
    N_{i,j} =
    \begin{cases}
      \frac{T_{i,j}}{|F|} & \text{if i $\neq$ j}\\
      1 - \frac{T_{i,j}}{|F|}(\sum_{k=0}^{j-1} T_{i,j} + \sum_{k=j+1}^{|F|} T_{i,j}) & \text{if i = j}
    \end{cases}
\end{equation}

This ranking is optimized in the following way. Initialize ranking list r $\in R^{|F|}$ such that $r^0 \in R^{|F|}$ where $\{x \in r^0: x = \frac{1}{|F|}\}$ and minimize $\sum_{j = 0}^{|F|} (r^i - r^{i-1})_{j}$ by a fixed number of iterations\\
where i is iteration number\\
\begin{equation}
    r^i =
    \begin{cases}
      (r^0)^T & \text{if i = 0}\\
      (r^{i-1})^T N & \text{else}
    \end{cases}
\end{equation}\\
The run time of the algorithm is $O(n^2k)$, where n is number of iterations and k is length of Markov chain.

\subsection{Baselines}
Baseline models are the models that were built to classify the Spotify dataset by training it against the complete balanced dataset. The metrics used to evaluate the performance of these models were Accuracy, F1 Score and feature importance 

\subsubsection{Support Vector Machine}
\begin{figure}[h!]
    \centering
    \includegraphics[width=8cm]{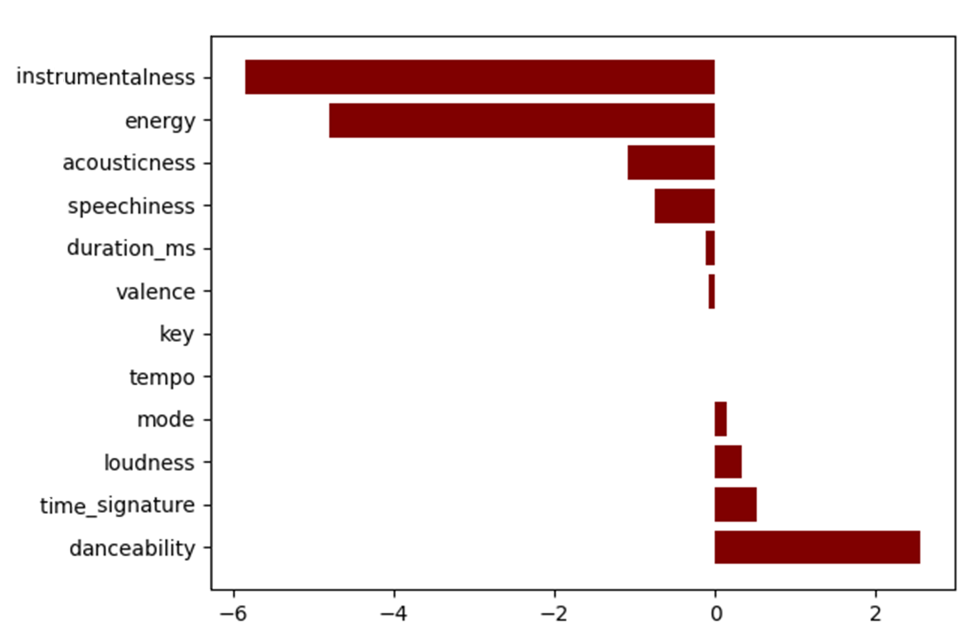}
    \caption{The feature importance values for the baseline SVM}
    \label{fig:featImpBaseSVM}
\end{figure}

\begin{figure}[h!]
    \centering
    \includegraphics[width=8cm]{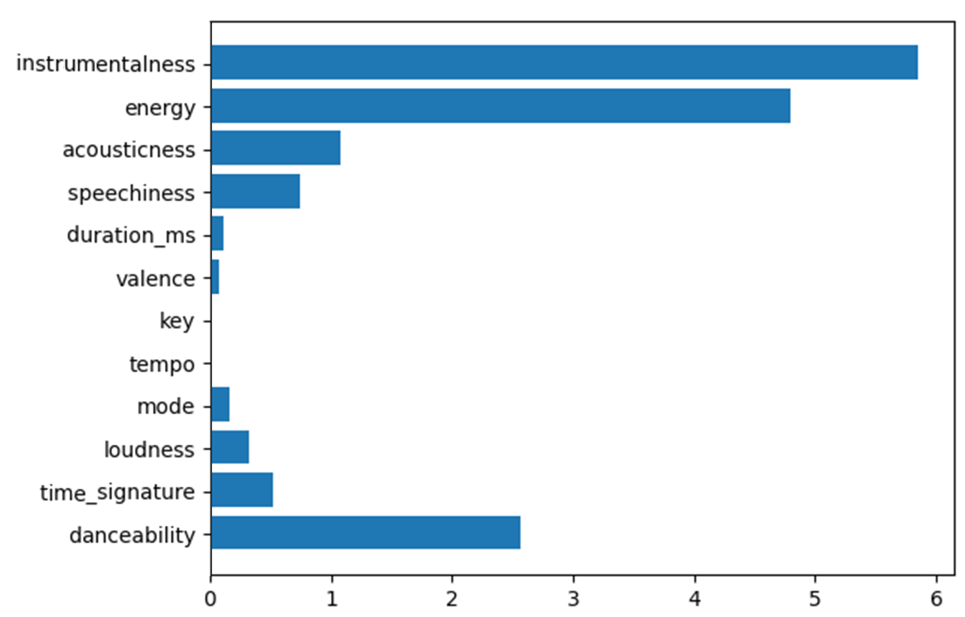}
    \caption{Absolute values of the weight vector for the baseline SVM}
    \label{fig:AbsBaseSVM}
\end{figure}

A Support Vector Machine is a linear classifier that partitions the data points into two classes by constructing a hyperplane in a d-dimensional space. The reasoning behind using a support vector machine is that with Linear Kernels, it is trivial to visualize how the model assigns importance to the features. The feature importance is simply the weight associated with each feature. The feature importance assigned by the baseline SVM model is shown in the figure ~\ref{fig:featImpBaseSVM}. It can be seen that features such as Instrumentalness and Acousticness have an inverse impact on the popularity of a song. Figure ~\ref{fig:AbsBaseSVM} shows the same features but with absolute values, giving essentially, a ranking of these features as seen by the SVM model on the balanced dataset. 

\subsection{Neural Network} 
\begin{figure*}[h!]
    \centering
    \includegraphics[width=15cm]{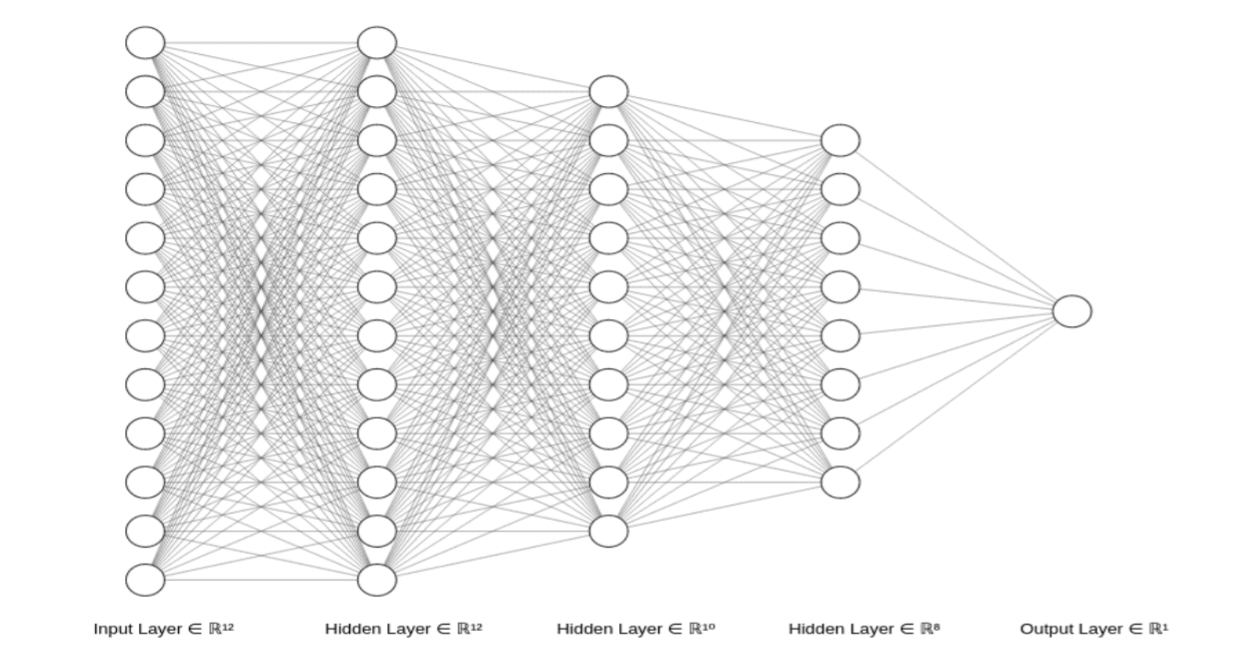}
    \caption{Neural Network Architecture}
    \label{fig:BaseNN}
\end{figure*}

A 3 layer neural network was constructed to classify the spotify dataset. The input layer has 12 neurons, for each of the 12 features of the dataset, that are learned. The first hidden layer is fully connected hidden layer with 12 neurons as well. The second hidden layer is fully connected and has 10 neurons. The third hidden layer has 8 neurons, and the output layer is a single neuron representing the binary class 0 and 1. For each of the hidden layers, we use the ReLU function as our activation function. For the output layer, we use the Sigmoid activation function. We avoid using dropout layers or other techniques that combat overfitting in order to have as simple of a baseline model as possible. The baseline neural network architecture is visualized in figure ~\ref{fig:BaseNN}. 

\subsubsection{Interpretability metrics}
For the baseline neural network, the following interpretability metrics were used.

\textit{Integrated Gradients} is an interpretability metric that is used to calculate the feature importance. It computes the gradient of the model's prediction to its input \cite{sundararajan2017axiomatic}. 
The approach is to first start with a baseline vector and construct new vectors until the model's predicted vector is reached. Gradients are computed at every step, for every input vector that is generated. 
For a feature i, the integrated gradient is defined as follows
\begin{equation}
    IG_x[i] = (x_i - x_i') \int_{\alpha=0}^{1} \frac{\partial F(x' +\alpha(x-x'))}{\partial x_i} d\alpha
\end{equation}

This feature was used to compute the importance of each neuron in the neural network. 

\textit{Neuron Conductance} is a technique to compute the conductance with respect to a particular neuron. It determines how important a neuron is in a given hidden layer \cite{dhamdhere2018important}. Conductance is denoted by this formula
\begin{equation}
    Cond_i^y(x) ::= (x_i - x_i') \int_{\alpha=0}^{1} \frac{\partial F(x' +\alpha(x-x'))}{\partial x_i} \frac{\partial y}{\partial x_i}d\alpha
\end{equation}

\begin{figure}[h!]
    \centering
    \includegraphics[width=9cm]{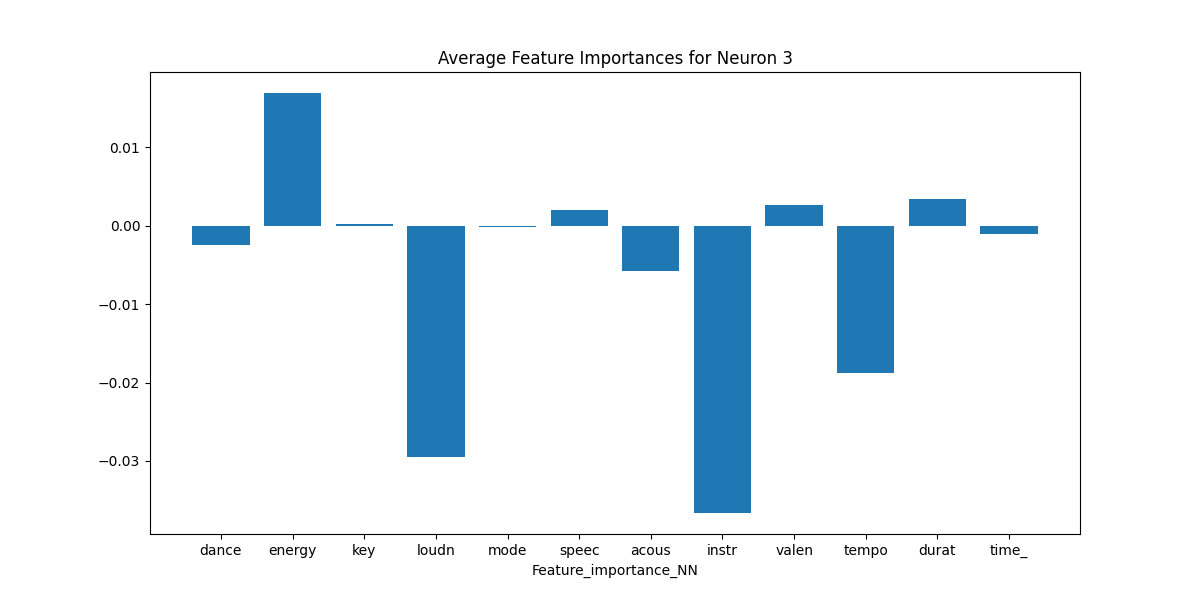}
    \caption{Feature importance calculated through IG for Layer 1 neuron 3}
    \label{fig:IGN3}
\end{figure}

Figure ~\ref{fig:IGN3} shows the feature importance calculated through IG method for neuron 3 in the first layer. 

Conductance for any layer is the difference in prediction between the layer and it's subsequent one, denoted by F(x) - F(x'). It follows the layer conservation principle, which states that the output is distributed through the layers into the input. 

\begin{figure}[h!]
    \centering
    \includegraphics[width=9cm]{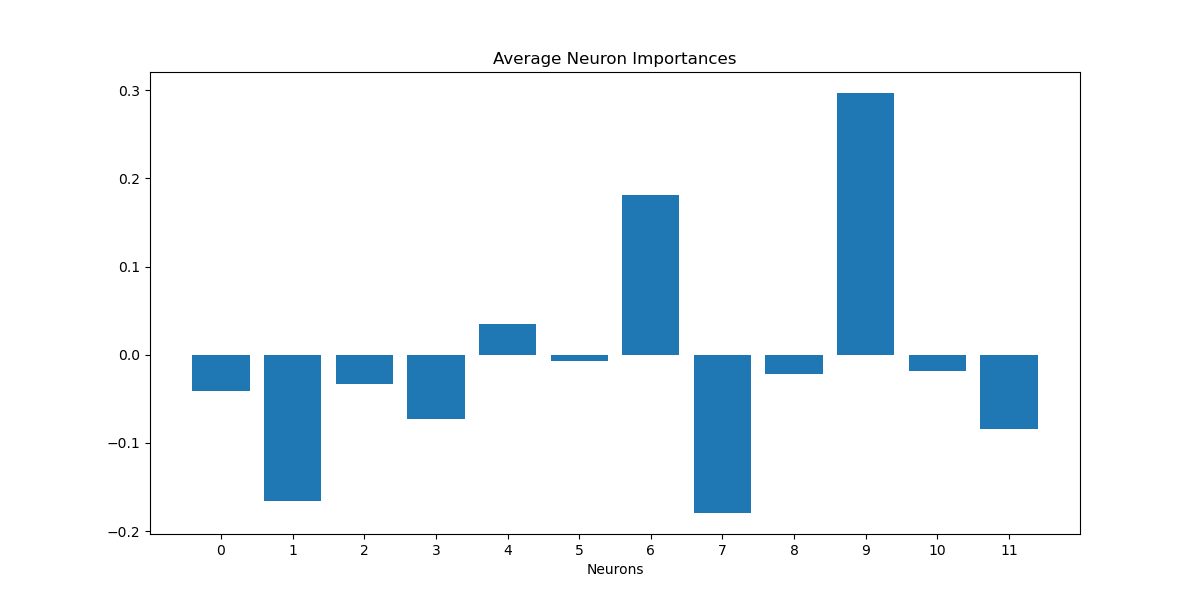}
    \caption{Neuron Importance calculated through Conductance}
    \label{fig:NeurCond}
\end{figure}

Figure ~\ref{fig:NeurCond} shows the importance of each neuron in the first layer. 

\textit{Average feature weights of a layer} is defined by us as the linear combination of neuron conductance and Integrated gradients. This provided us with a heuristic to determine how the layer treated the input features. We defined it in the following manner. 

\begin{equation}
    feature_imp(w) = \sum_{i \in L} Cond[i] * IG[i][w]
\end{equation}

\begin{figure}[h!]
    \centering
    \includegraphics[width=9cm]{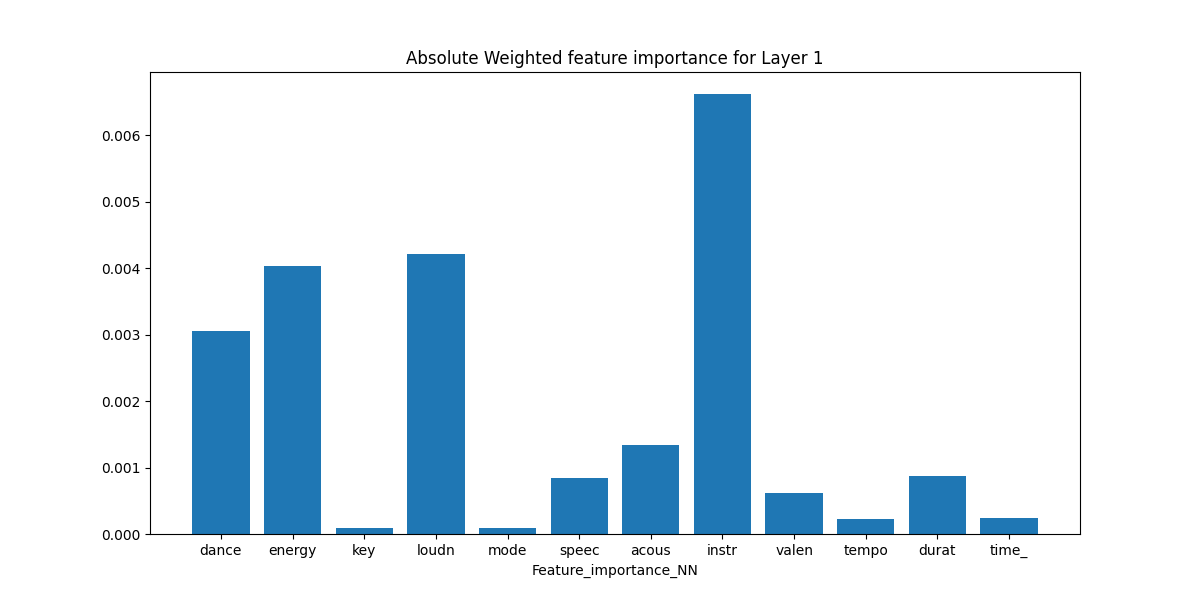}
    \caption{Weighted feature importance of Layer 1}
    \label{fig:weightedFeatImp}
\end{figure}

Figure ~\ref{fig:weightedFeatImp} shows weighted average feature importance of the first layer on the baseline model. 

\section{Implementation}
The two baseline models - SVM model and Neural Network model were trained on the fully balanced spotify dataset (80\% training, 20\% testing). The neural network was trained on varying sample sizes (500, 1000, 1500, 2000), biased true positive samples (20\%, 40\%, 60\%, 80\%), and starting weights (random weights, majority rank aggregated weights, MC4 rank aggregated weights, and Kemeney-Young rank aggregated weights). A 40\% Biased true positive sample refers to the fact that in the training data, only 40\% of the samples are positives and the other 60\% belong to the negative class. Since our dataset only has two target classes, we based our biased samples on the number of positive class labels they have (i.e 20\% true positive sample is a sample in which the target class = 1 is represented 20\% of the time). The SVM model was trained on a sample size of 500, with a biased true positive sample of 0.6. 
The varying sample sizes were done in order to try and see if the models performed better by being provided less a smaller biased sample. Furthermore, Since the neural network model is a very expressive model, we wanted to limit overfitting as best as we could by experimenting which sample sizes worked best.

The varying biased true positive samples were done to see how various starting weights of the models performed on biased samples of the dataset.
For weights, we scaled our rank aggregator results to values between \{-1, 1\}. The reason for doing this scaling is because Pytorch requires our weights to be between \{-1, 1\}. Our hope is that by providing weights that represent human intuition, we can bias the models to learn the distribution the sample comes from rather than the biased sample itself.

\section{Results and Discussion}

The table below shows the performance of the models and is followed by the discussion section. 

\begin{table*}[h!]
\begin{center}

\setlength{\tabcolsep}{10pt} % Default value: 6pt
\renewcommand{\arraystretch}{1.5} % Default value: 1
\begin{tabular}{ c  c  c }

\hline
 Model & Accuracy & F1 Score\\

 \hline
 \textbf{baseline models} & & \\
 SVM Model & 0.4898 & 0.0325\\
 Neural Network & 0.8570 & 0.8608 \\
 \hline
 \textbf{SVM Models} & & \\
 TP 0.6 SVM + human ranks & \textbf{0.5133} & \textbf{0.6750}\\
 TP 0.6 biased SVM without human ranks & 0.5328 & 0.2847\\
 \hline
 \textbf{Neural Network + MC\_4 Aggregator} & & \\
 TP 0.2 epoch 50 & 0.5820 & 0.3468 \\
 TP 0.2 epoch 200 & 0.6265 & 0.4653 \\
 TP 0.4 epoch 50 & 0.7570 & 0.7384\\
 TP 0.4 epoch 200 & 0.6977 & 0.6440 \\
 TP 0.6 epoch 50 & \textbf{0.7977} & \textbf{0.8159}\\
 TP 0.6 epoch 200 & 0.7594 & 0.7809\\
 TP 0.8 epoch 50 & 0.7344 & 0.7906\\
 TP 0.8 epoch 200 & 0.7203 & 0.7795\\
 \hline
 \textbf{Neural Network + Kemeny-Young Aggregator} & & \\
 TP 0.2 epoch 50 & 0.5648 & 0.2904 \\
 TP 0.2 epoch 200 & 0.5875 & 0.3561 \\
 TP 0.4 epoch 50 & 0.7555 & 0.7313 \\
 TP 0.4 epoch 200 & 0.7555 & 0.7354\\
 TP 0.6 epoch 50 & \textbf{0.7914} & \textbf{0.8221}\\
 TP 0.6 epoch 200 & 0.7695 & 0.8076\\
 TP 0.8 epoch 50 & 0.7313 & 0.7897\\
 TP 0.8 epoch 200 & 0.7313 & 0.7899\\
 \hline 
 \textbf{biased Neural Network (no human input)} & &\\
 TP 0.2 epoch 50 & 0.5758 & 0.3221\\
 TP 0.2 epoch 200 & 0.5867 & 0.3695\\
 TP 0.4 epoch 50 & 0.7406 & 0.7182\\
 TP 0.4 epoch 200 & 0.6938 & 0.6494\\
 TP 0.6 epoch 50 & \textbf{0.7953} & \textbf{0.8141}\\
 TP 0.6 epoch 200 & 0.7695 & 0.7944\\
 TP 0.8 epoch 50 & 0.7523 & 0.8013\\
 TP 0.8 epoch 200 & 0.7313 & 0.7923\\
 \hline

\end{tabular}

%\caption{Performance of the Models}
\label{table1}
\caption{Performance of the Models for Sample Size of 500}
\end{center}
\end{table*}

\subsection{Impact of human knowledge on SVM models}
The vanilla SVM model used a linear Kernel, making it easier to see the impact of human biased starting weights versus randomized starting weights. According to table 1, by setting random starting weights, we see that even on a completely balanced dataset, the SVM has an accuracy of just 48.98\%. This is due to us using a vanilla SVM with a linear Kernel. However, when we utilize human biased weights, we see that there is a slight increase in accuracy, even if the sample was biased, but a huge increase in the F1-score (0.6750). If we compare this F1 Score to the SVM that is trained on a biased sample but with randomized weights, we see that the F1-score is significantly smaller. Since our TP rate in the training set is 60\%, it is expected that the F1 score increases, since it is a metric that measures the accuracy in classifying positives. However, what is to be noted here is the increase in F1 score when human knowledge is utilized while training. 
\begin{figure}[h]
    \centering
    \includegraphics[width=9cm]{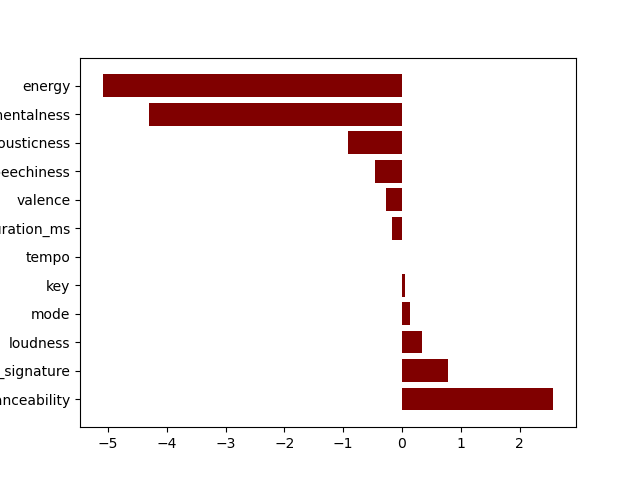}
    \caption{Feature importance of biased SVM}
    \label{fig:FeatImpBiasedSVM}
\end{figure}
We can see in Figure ~\ref{fig:FeatImpBiasedSVM} that the feature importance values are quite similar to the model trained without human knowledge.  Danceability still has the highest absolute value, while energy and instrumentalness have been swapped. This tells us that humans have a slightly different take on feature importance, but it is crucial that we capture this information, because this means that ML models are guided towards what humans consider to be important. 

\subsection{Impact of human knowledge on Neural Network models}
As seen in table 1, by utilizing either the MC\_4 Aggreagator or the Kemeny-Young Aggregator, we managed to increase both our accuracy and F1-score on the testing set for the neural network. This is especially evident for models with a true positive rate of 0.6 and trained for 50 epochs instead of 200. Another interesting note that we have found through our experimentation is that smaller biased sample sizes allow the model to generalize more. Similarly, by training for less epochs, we allow for better generalization of our neural network model. What this table demonstrates is that by providing human biased starting weights, we can improve the performance of an expressive model such as a neural network. 
The accuracy of the models increase as the ratio of positives increases. It reaches a peak at 50\%, which is to be expected. What this implies is that a fully balanced dataset leads to better learning. As we further increase the ratio of positives, the model becomes biased in favor of positives and the accuracy starts to drop. F1 score continues to increase since it simply measures the ability of the model to classify positives. 

However, what needs to be noted is that the use of human weights improves the performance considerably as opposed to training the neural network on biased data but with random initial weights. Specifically, when the true positive ratio is 40\%, the accuracy of the biased model trained on random weights is 69.38\%, while the models trained on human assigned initial weights have accuracy of 75.5\%, a considerable improvement. Similarly, in the same case, the F1 scores improved from 64.4\% to 73.54\%. 
It can be seen from the table that using human weights generally improved the performance when the training data was biased, with the only exception being the case when the true positive rate was 80\%, which implies that the data was too imbalanced.

\begin{figure}[h!]
    \centering
    \includegraphics[width=9cm]{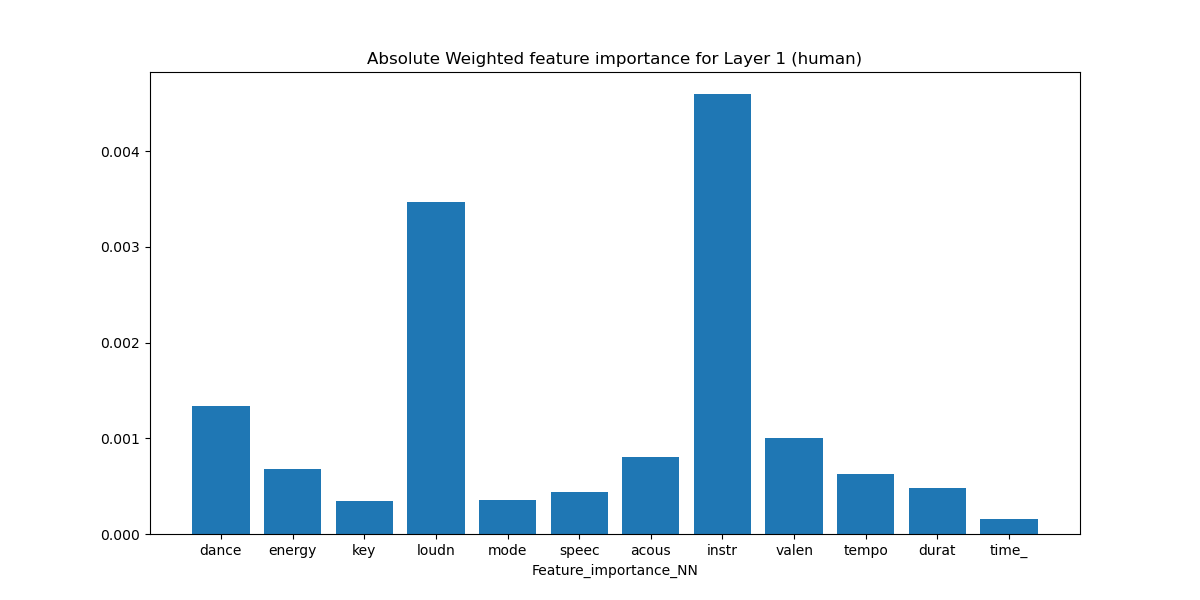}
    \caption{Human Absolute Weighted \newline Feature Importance for Layer 1}
    \label{fig:Abs}
\end{figure}

Figure ~\ref{fig:Abs} shows the absolute weights of the features as seen by the first layer of the neural network, that was trained on human weights. This netowork gives highest importance to instrumentalness and loudness, followed by danceability, which is significantly lower than the other two. The other attributes have lower weights, with time signature having the least weight. In the network trained on balanced data however, Instrumentalness and loudness were the features with the highest weights, but energy and danceability were comparable to loudness. Key was the attribute that had the least weight. 
\begin{figure}[h!]
    \centering
    \includegraphics[width=9cm]{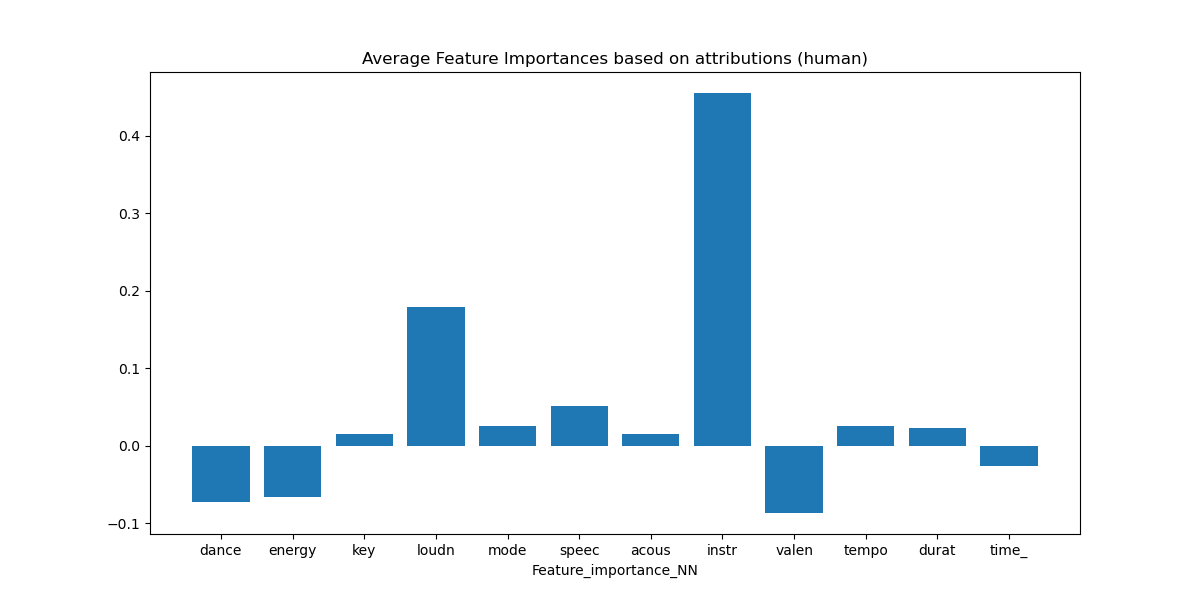}
    \caption{Average Feature Importance's based on Attributions}
    \label{fig:AvgFeatAttrib}
\end{figure}

Figure ~\ref{fig:AvgFeatAttrib} Shows the raw feature importance of first layer without considering neuron conductance. 

\begin{figure}[h!]
    \centering
    \includegraphics[width=9cm]{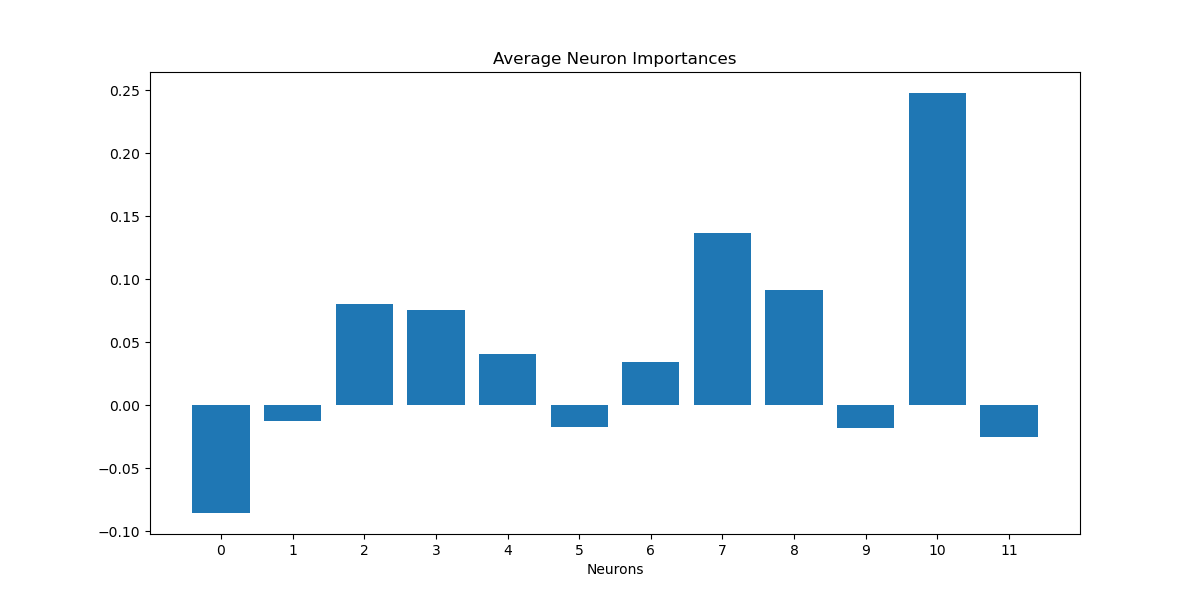}
    \caption{Average Neuron Conductance of first Layer}
    \label{fig:NeuronImpFirst}
\end{figure}

Figure ~\ref{fig:NeuronImpFirst} Shows the Conductance of the neurons of the first layer. These values vary significantly compared to the neuron conductance calculated on the model trained on balanced set. 

\begin{figure}[h!]
    \centering
    \includegraphics[width=9cm]{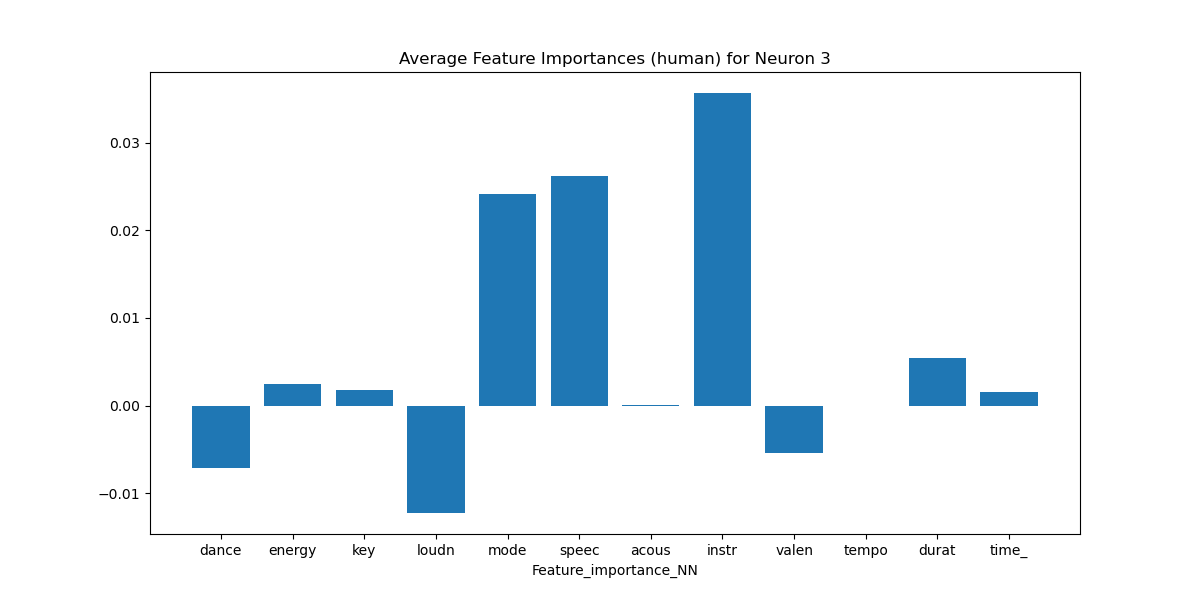}
    \caption{Average Feature Importance's (human) for Neuron 3}
    \label{fig:Neur3FeatImp}
\end{figure}

Figure ~\ref{fig:Neur3FeatImp} also varies significantly compared to the feature importance calculated on the third neuron, when trained on the balanced set, suggesting that using human weights indeed has considerable effect on both Integrated Gradient and Neuron Conductance. 

\section{Future Work}
This work tested the use of prior human knowledge as initial weights for the model. While this may work for tabular data in binary classification settings, work needs to be done in determining how human knowledge can be incoporated into settings that involve multi-class classification, regression or in the case of Language models. 
Aggregating opinion of users by considering merely the general knowledge of the humans might not be sufficient while attempting to solve problems which need a far more sophesticated understanding of the subject. For, instance, while classifying medical images, domain experts cannot be eliminated and their opinions are more important than the opinions of someone who has a basic understanding of the topic. In such cases, work needs to be done in evaluating the relative worth of expert opinions based on expertise. 

Further work is required in determining whether adding the human component into the training process improves the fairness of the model and care needs to be taken to check whether the human opinions themselves are not negatively biased. The humans involved in providing opinions need to be fully representative of the society when trying to solve more social problems with machine learning. 
Work needs to be done to determine if there are other practical and applied ways to capture human domain knowledge and integrate it into ML frameworks. 

Furthermore, we need to investigate how features may be correlated with each other. We need to consider scenarios where multiple features, when combined, provide a better understanding of the datapoint, as opposed to them individually. We also need to account for the presence of external confounding factors which may affect the datapoint, but such data may not be available in the training dataset. It needs to be proved, whether humans capture such confounding information, when the provide their opinions.

% Use \bibliography{yourbibfile} instead or the References section will not appear in your paper
\bibliography{aaai22}

\end{document}